\documentclass[12pt]{iopart}

\usepackage{graphicx}
\usepackage{moreverb,url}
\expandafter\let\csname equation*\endcsname\relax
\expandafter\let\csname endequation*\endcsname\relax

\usepackage{amsmath}
\begin{document}

\title{A general locomotion control framework for multi-legged locomotors}

\author{Baxi Chong$^{1*}$, Yasemin O. Aydin$^{1*}$, Jennifer M. Rieser$^{2}$, Guillaume Sartoretti$^{3}$, Tianyu Wang$^{1}$, Julian Whitman$^{4}$, Abdul Kaba$^{5}$, Enes Aydin$^{1}$, Ciera McFarland$^{6}$, Kelimar Diaz Cruz$^{1}$, Jeffery W. Rankin$^{7}$, Krijn B  Michel$^{8}$, Alfredo Nicieza$^{9}$, John R Hutchinson$^{8}$, Howie Choset$^{4}$ and Daniel I. Goldman$^{1}$}

\address{$^{1}$ Georgia Institute of Technology 
$^{2}$ Emory University  
$^{3}$ National University of Singapore
$^{4}$ Carnegie Mellon University
$^{5}$ Morehouse College \
$^{6}$ Pennsylvania State University \
$^{7}$ Rancho Research Institute,
$^{8}$ Royal Veterinary College,
$^{9}$ University of Oviedo,
$^{*}$ Equal contribution
}
\ead{daniel.goldman@physics.gatech.edu}
\vspace{10pt}

\begin{indented}
\item[Jan 2022]
\end{indented}

\begin{abstract}
Serially connected robots are promising candidates for performing tasks in confined spaces such as search and rescue in large-scale disasters. Such robots are typically limbless, and we hypothesize that the addition of limbs could improve mobility. However, a challenge in designing and controlling such devices lies in the coordination of high-dimensional redundant modules in a way that improves mobility. Here we develop a general framework to discover templates to control serially connected multi-legged robots. Specifically, we combine two approaches to build a general shape control scheme which can provide baseline patterns of self-deformation (``gaits'') for effective locomotion in diverse robot morphologies. First, we take inspiration from a dimensionality reduction and a biological gait classification scheme to generate cyclic patterns of body deformation and foot lifting/lowering, which facilitate generation of arbitrary substrate contact patterns. Second, we extend geometric mechanics, which was originally introduced to study swimming in low Reynolds number, to frictional environments, allowing identification of optimal body-leg coordination in this common terradynamic regime. Our scheme allows the development of effective gaits on flat terrain with diverse number of limbs (4, 6, 16, and even 0 limbs) and backbone actuation. By properly coordinating the body undulation and leg placement, our framework combines the advantages of both limbless robots (modularity and narrow profile) and legged robots (mobility). Our framework can provide general control schemes for the rapid deployment of general multi-legged robots, paving the way toward machines that can traverse complex environments. In addition, we show that our framework can also offer insights into body-leg coordination in living systems, such as salamanders and centipedes, from a biomechanical perspective.

\end{abstract}
\maketitle

\section*{Introduction}

Robots with different numbers of limbs have different advantages. Quadrupeds are known for their agility \cite{tan2018sim}, whereas hexapods and myriapods for their stability \cite{aoi2016advantage,saranli2001rhex}, and limbless robots for their ability to fit into confined spaces \cite{mckenna2008toroidal}. 
But robots with increasing complexity and numbers of degrees-of-freedom (DoF) present challenges in motion coordination, which if not addressed, may render them unusable.
Furthermore, the diversity of shape and form makes it challenging to transfer control insights gained from one platform onto another. 
We are left with limited intuition and physical understanding of how to coordinate the many DoF in diverse and complex robots to generate effective locomotion.
    
To address the growing need to control robots with different shapes, modular robot control strikes a balance between encompassing a  variety of shapes while still being able to precisely control them \cite{wright2007design,tesch2009parameterized}. Modular robot control has been successfully used in serially connected limbless robots where a single control principle can be applied in robots with different sizes \cite{sartoretti2019distributed}. In contrast, the study of modular control in general multi-legged robots has been limited. The challenge in serially connected multi-legged robots lies not only in designing the stepping patterns of legs, but also in the coordination between the body and legs. For example, in robots that combine limbs and body undulation, if stepping patterns and body undulations are not properly coordinated, limbs can interfere with each other, resulting in reduced locomotor performance, instability, or even failure \cite{mazouchova2013flipper,byl2008metastable}.

We would like to develop control schemes to generate effective periodic ``self-deformation patterns"\footnote{We consider self-deformation patterns as relative movement of body and limb elements} for the general class of serially connected legged and limbless robots.
Over the past decades, many techniques (e.g., gait generation \cite{wettergreen1992gait,sun2001automating}, central pattern generators \cite{ijspeert2007swimming,crespi2008controlling}, nearest limb synchronization \cite{dutta2019programmable}, and learning methods \cite{tan2018sim,hwangbo2019learning}) have been developed, each of which can control some specific robot type  \cite{saranli2001rhex,park2017high,hatton2013geometric,marvi2014sidewinding,ijspeert2007swimming}.
In this paper, we take inspiration from living systems: organisms with diverse numbers of appendages and body plans exhibit effective locomotion on almost all terrestrial environments~\cite{farley1997mechanics,popovic2005ground,marvi2014sidewinding} by making/breaking the ground contact with limbs (e.g., salamanders) and bodies (e.g., sidewinders) in conjunction with waves of undulation. 

One method used over the last century to understand legged locomotion is a classification scheme called ``Hildebrand diagrams". In 1965, Hildebrand  \cite{hildebrand1965symmetrical} developed schemes to study symmetric gaits\footnote{In symmetric gaits, the contralateral (left and right pair) of legs are $180^\circ$ out of phase.} observed in quadrupedal animals (e.g., horses).
These gaits have two key variables: \emph{duty factor}, the fraction of a period that each leg is on the ground over a full gait cycle, and \emph{lateral phase lag}, the fraction of a period that the hind leg leads the foreleg on the same side. Both key variables are modulated in response to speed changes in biological \cite{hildebrand1980adaptive,white1994locomotor,hutchinson2019divergent}. 
Using these gait principles as a reference, a multitude of algorithms have been developed for quadrupedal robot locomotion or to explain why living quadrupeds choose certain gaits  \cite{mcghee1968stability,bai1999new,owaki2013simple,kalakrishnan2010fast,bellicoso2018dynamic,hoyt1981gait}. But thus far, these gait principles have not been applied to robots with more than four appendages. For multi-legged robots, there is a lack of a systematic gait description framework that allows us to modulate the balance between locomotion metrics such as speed and stability.

In animals and increasingly in robots, appendages that make direct contact with substrates are not the sole contributor to locomotion. Undulatory body motions play an important role in generating propulsive forces in many systems~\cite{kafkafi1998traveling,farley1997mechanics,frolich1992kinematic,manton1952evolution}. For undulatory locomotors, the geometric mechanics community ~\cite{kelly1995geometric,ostrowski1998geometric,shapere1989geometric,shammas2007geometric,hatton2015nonconservativity,astley2015modulation,chong2021coordination,toruspaper} has developed a gait design framework to prescribe self-deformations of systems immersed on continuous media, such as 3-link robots, lizards, and snakes~\cite{rieser2019geometric,hatton2015nonconservativity,hatton2013geometric}; and in discontinuous settings, including sidewinders \cite{astley2020surprising,astley2020side}. While mathematically elegant, geometric mechanics has limitations. In particular, it is not directly applicable to  systems with a large number of appendages. Furthermore, despite some recent efforts~\cite{rieser2019geometric,chong2021frequency}, application of geometric mechanics in frictional environments (e.g., rate-independent Coulomb dry friction) has not been systematically studied. Therefore, we must develop dimensionality reduction and physical modelling methods before we can use geometric mechanics to design gaits for serially connected multi-legged robots. 

\begin{figure}
    \centering
    \includegraphics[width=0.8\linewidth]{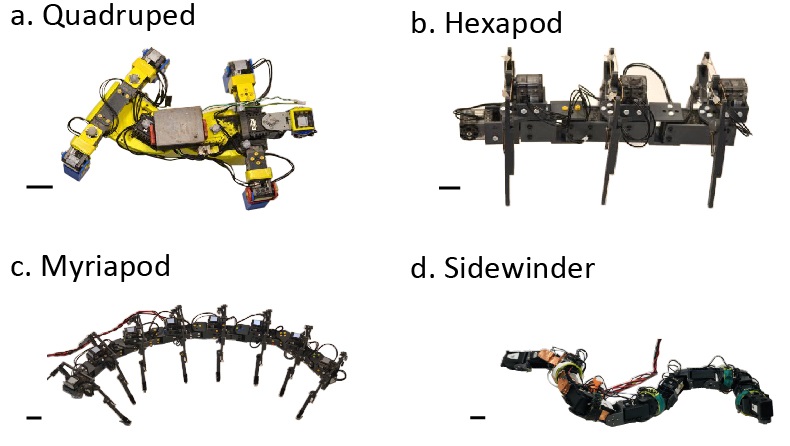}
    \caption{\textbf{Legged and limbless robotic models studied in the paper.} a. Quadrupedal robot \cite{ozkanAydin_LivingMach,chong2021coordination} b. Hexapod robot c. Myriapod robot with eight pairs of legs \cite{OzkanAydinRobosoft2020}  d. Sidewinder robot \cite{Astley_2020}. All scale bars are 5 cm. See Fig. S4 for the axis of joint angles.}
    \label{fig:allrobot}
    \vspace{-2em}
\end{figure}

In this paper, we integrate dimensionality reduction techniques with tools from geometric mechanics to develop locomotion control schemes for serially connected robots. We first extend the Hildebrand gait classification scheme to prescribe a wide range of contact patterns (the sequence of making/breaking contacts with environments) using the classical Hildebrand parameters (duty factor and lateral phase lag).
We use the extended Hildebrand scheme to reduce dimensionality and prescribe body undulation as a traveling wave. In doing so, we can apply geometric mechanics to coordinate the lateral body undulation and the limb contact patterns. 
We evaluate gait performance based on speed and static stability, and investigate the relationship between these metrics and the Hildebrand parameters. We demonstrate our motion control framework on robots with four (quadrupedal), six (hexapod), 16 (myriapod-like), and even zero (snake-like) limbs (Fig. \ref{fig:allrobot}). 
Our analysis reveals empirical rules to balance the trade-off between speed and static stability, and the potential benefit of body undulation in multi-legged robot locomotion. 

Moreover, by properly coordinating lateral body undulation and leg movement, our framework leverages advantages from both legged and limbless robots. Specifically, our framework facilitates centralized control of serially connected multi-legged robots by introducing waves in both limb contact and lateral body undulation. With properly coordinated lifting and landing body segments, our framework can also improve the mobility of limbless robots by giving insights into coordination and trade-offs of stability and speed in serially connected multi-legged robots. In this way, our framework offers the potential to modulate gaits for different tasks by switching between fast gaits and stable gaits. Further, we show that our scheme can generate control hypotheses for diverse living systems including salamanders and centipedes, thereby offering new insights on the functional role of body-leg coordination from a biomechanical and robophysical perspective.


\section*{Hildebrand Gait Prescription}

\subsection*{Related work}

In the Hildebrand gait formulation \cite{hildebrand1965symmetrical}, symmetric quadrupedal gaits are categorized by two parameters: The duty factor represents the fraction over a gait period that each leg is on the ground, and the lateral phase lag represents the fraction over a gait period that the hind leg leads the foreleg on the same side. 
There are three major assumptions in the Hildebrand symmetric gait family: (1) the duty factor of each leg is the same, (2) the pairs of contralateral legs are $180^\circ$ out of phase, and (3) the lateral phase lag is the same for left and right legs.

\begin{figure*}
    \centering
    \includegraphics[width=1\linewidth]{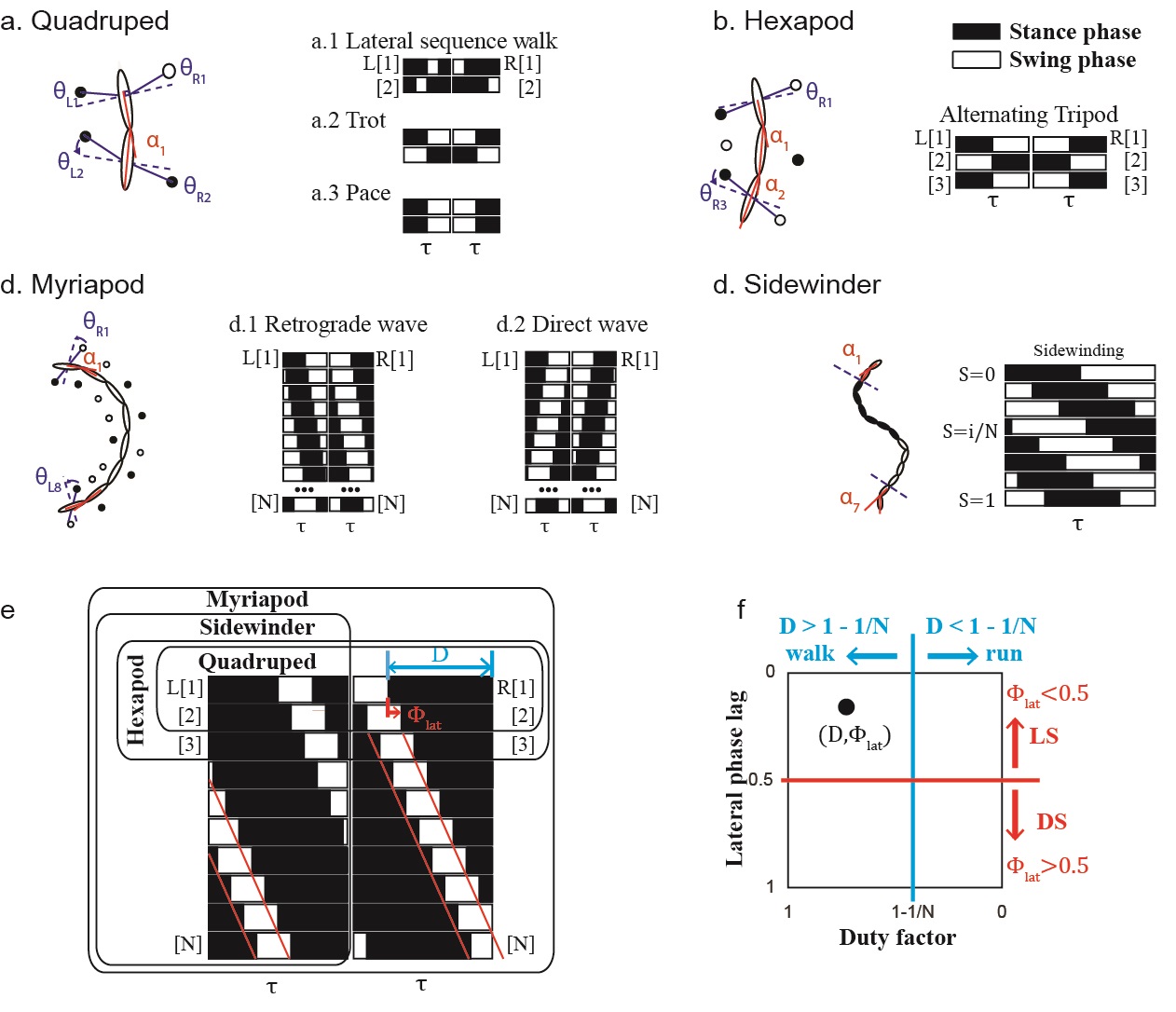}
    \caption{\textbf{Modelling multi-legged systems and sidewinders.}  The contact patterns of some well-known gaits: (a.1) lateral sequence walking, (a.2) trotting, and (a.3) pacing in quadrupeds; (b) alternating tripod in hexapods, (c) sidewinding in snake-like limbless robots, (d.1) retrograde-wave and (d.2) direct-wave gaits in myriapods. For each system, these diagrams show the variables included in the model, such as leg joint angles $\theta_N$, and body joint angles $\alpha_{(N-1)}$, where $N$ is the number of leg pairs for legged systems or joint sets in the sidewinder. In the contact sequence diagrams, filled blocks represent stance phase, and open blocks represent swing phase. (e) A general contact pattern table. The blue arrow represents the duty factor $D$. The red arrow represents the lateral phase lag, $\Phi_{lat}$. $\tau$ denote gait phase. (f) Hildebrand plots with two parameters $D$ and $\Phi_{lat}$ to characterize the motions in the vertical plane. We labeled the region associated with walking, running, lateral sequence (LS) and diagonal sequence (DS) gaits.
 }
    \label{fig:prescribe}
\end{figure*}

We use a binary variable $c$ to represent the contact state of a leg, where $c=1$ represents the stance phase and $c=0$ represents the swing phase. The contact pattern of symmetric quadrupedal gaits can be written as
\begin{align}
    c_{FL}(\phi_c) &=\begin{cases}
      1, & \text{if}\ \text{mod}(\phi_c,2\pi) < 2\pi D\\
      0, & \text{otherwise}
    \end{cases} \nonumber\\
    c_{FR}(\phi_c) &= c_{FL}(\phi_c+\pi) \nonumber\\
    c_{HL}(\phi_c) &= c_{FL}(\phi_c+2\pi \Phi_{lat}) \nonumber\\
    c_{HR}(\phi_c) &= c_{FL}(\phi_c+2\pi \Phi_{lat} +\pi) \label{eq:hilde_quad}
\end{align}

\noindent where $\Phi_{lat}$ denotes the lateral phase lag, $D$ the duty factor, $ c_{FL}(\phi_c)$, $ c_{FR}(\phi_c)$, $ c_{HL}(\phi_c)$, and $ c_{HR}(\phi_c)$ the contact state of the fore-right (FR), fore-left (FL), hind-left (HL), and hind-right (HR) limbs at the gait phase $\phi_c$, respectively. 
Many common quadrupedal gaits can be described using the Hildebrand formula. 
For example, the lateral sequence walking gait (Fig. \ref{fig:prescribe}) can be described by $D=0.75$, $\Phi_{lat}=0.25$. Plotting a diagram of the stance/swing phases of the feet from just these two parameters shows that in this gait, each leg is lifted for a quarter of a cycle and only one leg is lifted at any given instant, and the leg lifting sequence follows FR, HR, FL, and HR. 
The trot gait (Fig. \ref{fig:prescribe}) can be described by $D=0.5$, $\Phi_{lat}=0.5$, where the FR and HL are coupled in phase (same as the FL and HR pair). 
Another quadrupedal gait, the pace gait (Fig. \ref{fig:prescribe}), can be described by $D=0.5$, $\Phi_{lat}=0$, where the FR and HR are coupled in phase, as are the FL and HL pair. Note that asymmetric quadrupedal gaits, such as bounding and galloping, exist and cannot be prescribed by the same Hildebrand methods.

\subsection*{Prescription of Contact Patterns for Arbitrary Robots}

The first two assumptions of the Hildebrand symmetric gait family can hold in general for non-quadrupedal systems with discrete contacts. To expand the third assumption to a broader range of locomotors, we can generalize the definition of the lateral phase lag to be the phase lag between two consecutive legs (instead of only the fore and hind legs) on the same side.
Then, the contact function of a multi-legged system can be written as:
\begin{align}
     c_l(\phi_c,1) &=\begin{cases}
      1, & \text{if}\ \text{mod}(\phi_c,2\pi) < 2\pi D\\
      0, & \text{otherwise}
    \end{cases} \nonumber \\
    c_l(\phi_c,i) &= c_l(\phi_c + 2\pi(i-1) \Phi_{lat},1) \nonumber \\
    c_r(\phi_c,i) &= c_l(\phi_c+\pi,i),   
    \label{eq:general_hild}
\end{align}

\noindent where $c_l(\phi_c,i)$ (and $c_r(\phi_c,i)$) denotes the contact state of $i$-th leg on the left (and the right) at gait phase $\phi_c$, $i\in \{1, ... N\}$ for $2N$-legged systems. 

Many common multi-legged gaits can also be described by this extended Hildebrand formulation.
For example, many hexapod robots and animals use the alternating tripod gait (Fig. \ref{fig:prescribe}), which couples FL, MR (middle-right), and HL in phase, and couples the FR, ML, and HR similarly. 
The alternating tripod gait for a hexapod ($N=3$) can be described by $D=0.5$ and $\Phi_{lat}=0.5$.

Myriapod gaits can be classified into direct waves and retrograde waves of limb contact \cite{kuroda2014common} (Fig. \ref{fig:prescribe}). 
Typically, for gaits with $\Phi_{lat}<0.5$, the phase of the hind leg is ahead of the phase of its immediate fore leg. In other words, the legs move in a wave propagating from tail to head, which we call a diagonal sequence gait, and which corresponds to direct waves in myriapods. On the other hand, when $\Phi_{lat}>0.5$, the phase of the hind leg is behind the phase of its immediate fore leg. Therefore, the leg wave propagates from head to tail, which we call a lateral sequence gait, and which corresponds to retrograde waves in myriapods. 
Interestingly, on level ground, animals with fewer legs more commonly use lateral sequence gaits \cite{hildebrand1965symmetrical,hildebrand1967symmetrical,hildebrand1980adaptive,sukhanov1974general}, and animals with more legs use both diagonal sequence and lateral sequence gaits~\cite{manton1952evolution,anderson1995axial}.  
As we will discuss later, we hypothesize that this difference in gait choice is a result of the balance between speed and stability.

Our proposed gait formulation can also include systems without legs, e.g., sidewinding limbless robots. 
The complex mode of limbless locomotion, sidewinding, can be prescribed as the superposition of two waves: lateral and vertical body waves~\cite{astley2015modulation}.
Similar to legged systems, sidewinders can regulate their contacts by modulating the vertical traveling wave \cite{astley2015modulation}. 
The typical contact pattern of a sidewinder is shown in Fig. \ref{fig:prescribe}. 
Note that the contact pattern during sidewinding locomotion is the same as one side (either left or right) of the contact pattern of a legged system. 
As such, we prescribe the contact state of the $i$-th link of the sidewinding system as $c(\phi_c,i)=c_l(\phi_c,i)$, where $c_l(\phi_c,i)$ is defined in Eq. \ref{eq:general_hild}

\subsection*{Prescription of Leg Shoulder Movement}

Legs generate self-propulsion by protracting during the stance phase to make contact with the environment, and retracting during the swing phase to break contact. 
That is, the leg moves from the anterior to the posterior end during the stance phase and moves from the posterior to anterior end during the swing phase. 
With this in mind, we use a piece-wise sinusoidal function to prescribe the anterior/posterior excursion angles ($\theta$, Fig. \ref{fig:prescribe}) for a given contact phase ($\phi_c$),
\begin{align}
        \theta_l(\phi_c,1)  &=\begin{cases}
      A_{\theta}\cos{(\frac{\phi_c}{2D})}, & \text{if}\ \text{mod}(\phi_c,2\pi)  < 2\pi D\\
      -A_{\theta}\cos{(\frac{\phi_c-2\pi D}{2(1-D)})}, & \text{otherwise},
    \end{cases} \nonumber \\
    \theta_l(\phi_c, i) &= \theta_l(\phi_c + 2\pi (i-j)\Phi_{lat}, j) \nonumber \\
    \theta_r(\phi_c, i) &= \theta_l(\phi_c + \pi, i) 
    \label{eq:legmove}
\end{align}

\noindent where $A_\theta$ is the shoulder angle amplitude, $\theta_l(\phi_c,i)$ and $\theta_r(\phi_c,i)$ denote the leg shoulder angle of $i$-th left and right leg at contact phase $\phi_c$, respectively.  Note that the shoulder angle is maximum ($\theta=A_\theta$) at the transition from swing to stance phase, and is minimum ($\theta=-A_\theta$) at the transition from stance to swing phase.
Fig. \ref{fig:example_pre} shows an example of a hexapod gait under this equation.

\subsection*{Numerical Prediction on Speed and Stability}

\begin{figure}
    \centering
    \includegraphics[width=0.5\linewidth]{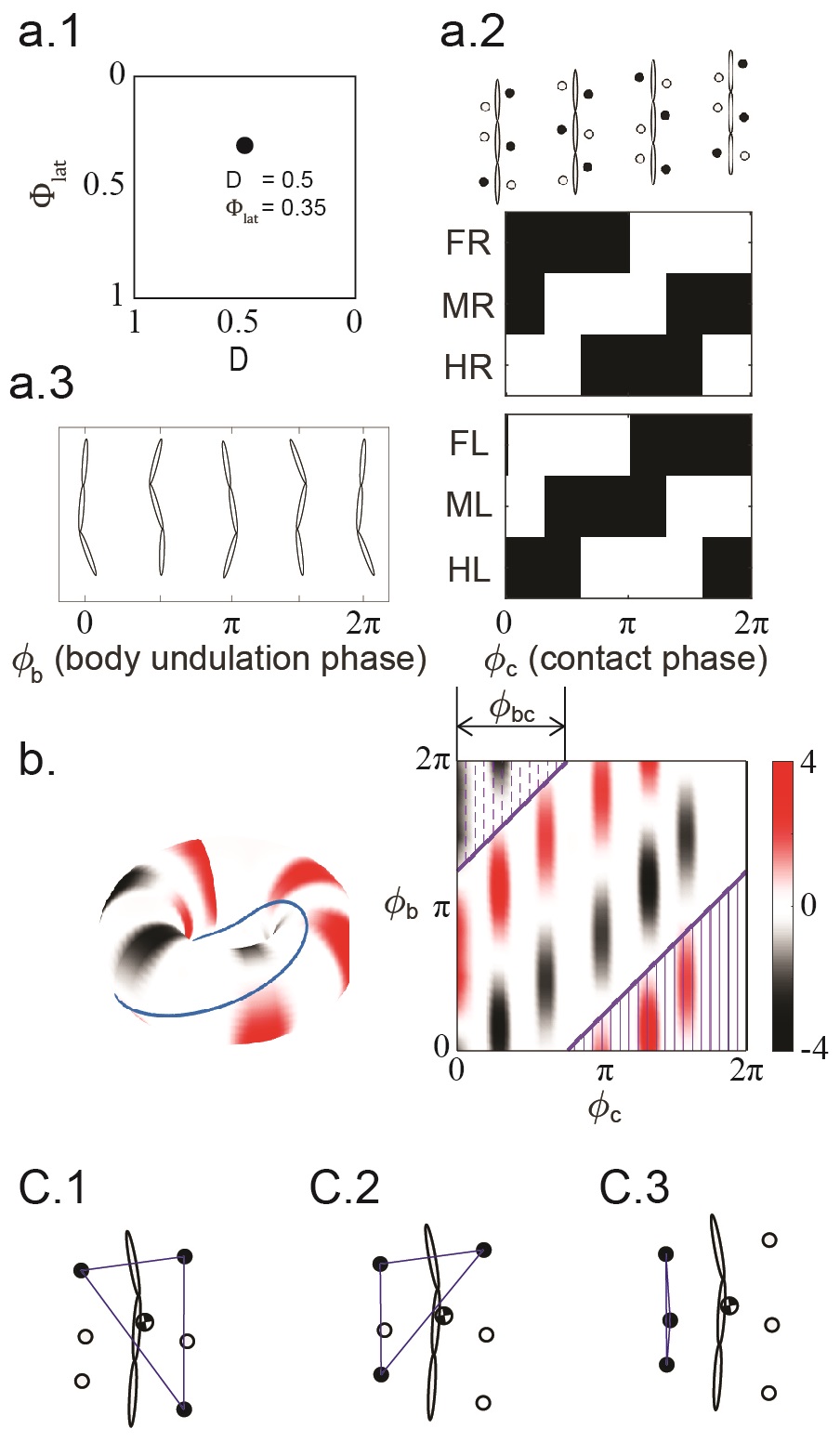}
    \caption{\textbf{An example of gait design for a hexapod using Hildebrand gait principles and geometric mechanics.} From the parameter space (a.1), we select the duty factor $D$ and lateral phase lag $\Phi_{lat}$. We prescribe the contact by its phase $\phi_c$ (a.2), and the lateral body undulation by its phase $\phi_b$ (a.3). (b) The gait parameters determine the equations of motion, which in turn are used to derive a height function, and design a gait. The gait path (the purple curve) shown maximizes the volume enclosed in the lower right corner (in solid shadow) minus the volume enclosed in the upper left corner (in dashed shadow). The left panel is the toroidal visualization of the  height function, the right panel is the Euclidean visualization of the  height function. Fig. c.1-3 illustrates configurations in which the robot is statically stable (c.1), statically unstable (c.2) and unstable (c.3)
 }
    \label{fig:example_pre}
\end{figure}

\begin{figure}
     \centering
     \includegraphics[width=0.8\linewidth]{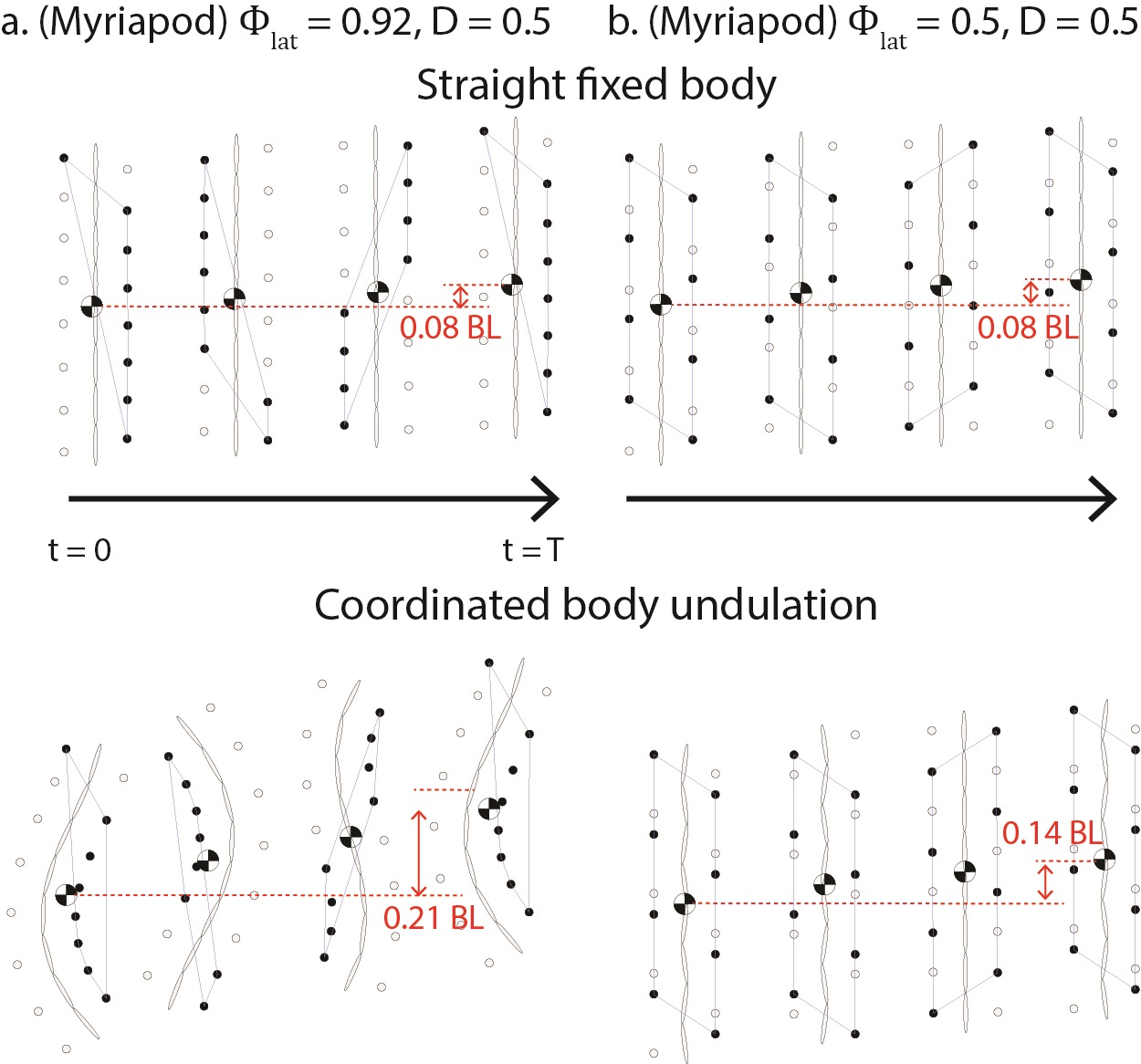}
     \caption{\textbf{Snapshots of the numerical simulation showing examples of two prescribed myriapod gaits.} (a) statically unstable, $\Phi_{lat}=0.92$, $D=0.5$ and (b) statically stable $\Phi_{lat}=0.5$, $D=0.5$. We compared the gait with straight fixed body (top) and gaits with coordinated body undulation (bottom). The displacement, in body length (BL) per cycle, are labeled with a red arrow. The black/white circles show the stance/swing phase of the feet. } 
     \label{fig:example_evaluate}
 \end{figure}

We numerically calculated the speed of various gaits \cite{hatton2015nonconservativity} over a range duty factors and lateral phase lags for a quadrupedal, hexapod, myriapod, and sidewinder systems. Fig. \ref{fig:example_pre} and Fig. \ref{fig:example_evaluate} graphically depict the process, and the Materials and Methods section provide details. 
To explicitly show the effect of limb-substrate contact on speed, we fixed the swing angle $A_\theta$ when comparing the displacements of different gait parameters.
Note that in this section, there is no body undulation in all gaits.

\begin{figure*}
    \centering
    \includegraphics[width=0.85\linewidth]{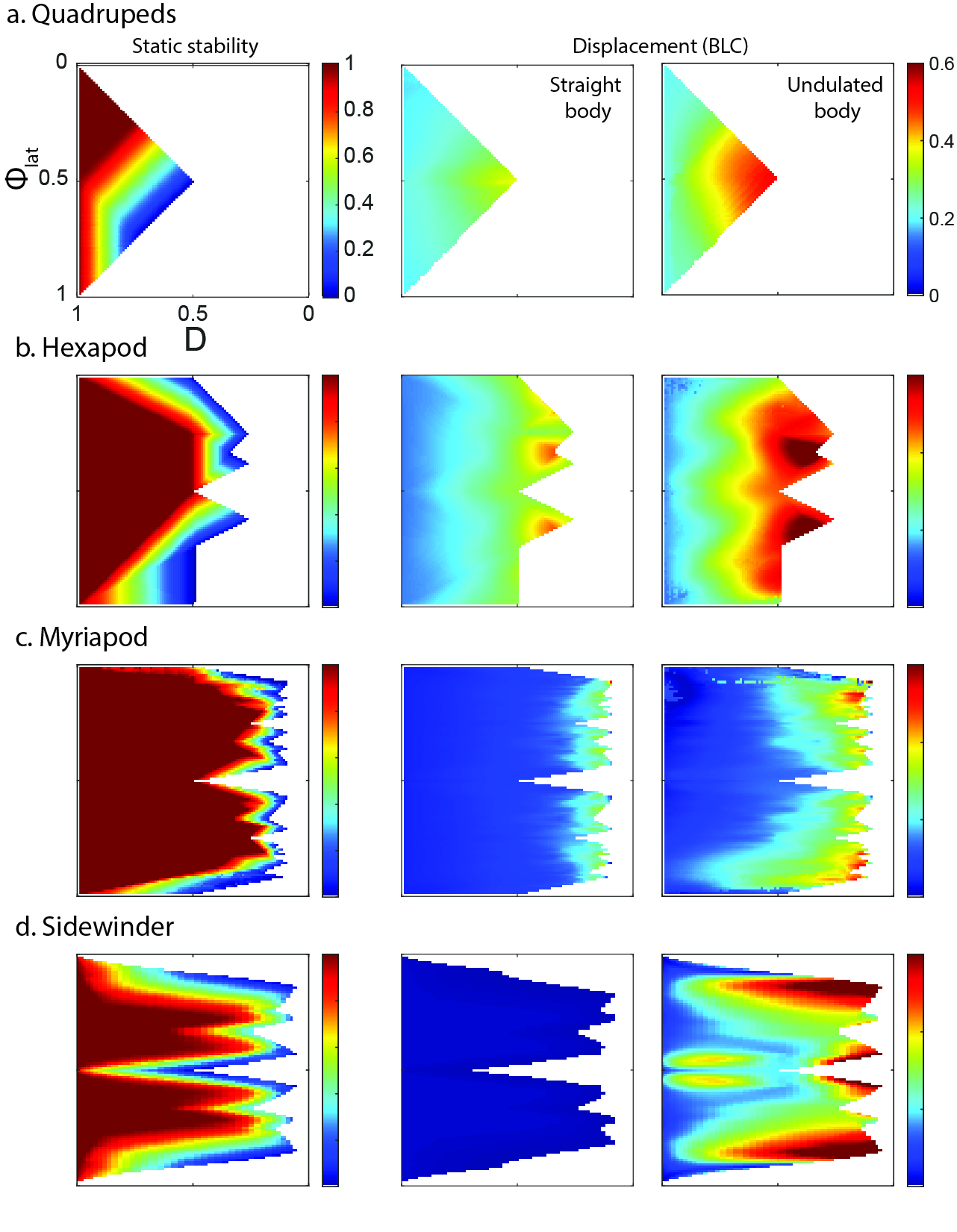}
    \caption{\textbf{Trade-off between speed and static stability in quadruped, hexapod, myriapod, and sidewinding systems.} Theoretically predicted static stability (left column), displacement in body lengths per cycle (BLC) with fixed straight back (middle column), and displacement  with coordinated lateral body undulation (right column) over the space of Hildebrand parameters $D$ and $\Phi_{lat}$, for the quadruped (a), hexapod (b), myriapod (c) and sidewinder (d). White space in all panels represents the regions where unstable configurations exist (Fig. \ref{fig:example_pre}c.3); we defined static stability to be zero in those regions. Note that static stability of the quadruped, hexapod and myriapod is numerically calculated for configurations with a straight backbone. The static stability of the sidewinder is numerically calculated for gaits with coordinated lateral body undulation. Note that we only consider gaits where unstable configurations (Fig. \ref{fig:example_pre}c.3) do not occur.}
    \label{fig:sim}
\end{figure*}

The numerical prediction of body speed, measured in units of body length per cycle (BLC), is plotted in Fig. \ref{fig:sim} (middle column).
We observe that modulating the lateral phase lag does not significantly affect body speed. 
This observation becomes more apparent for systems with more legs. In the myriapod system, speed is almost independent of the lateral phase lag and is uniquely determined by the duty factor.

In addition to measuring body speed, we require other metrics to quantify gait stability. For instance, the contact pattern of quasi-static gaits (e.g., quadrupedal walking gaits) need significantly less low-level control efforts to be stably realized on robots than the contact pattern of dynamically stable gaits (e.g., bouncing gaits) \cite{mcghee1968stability}. 
In this paper, we separate robots' configurations into three groups (1) statically stable, (2) statically unstable, and (3) unstable. 
In the statically stable configurations, the center of mass is bounded within the supporting polygon (Fig. \ref{fig:example_pre}c.1). 
In the statically unstable configurations, also known as unstable diagonal-couplet gaits \cite{hildebrand1967symmetrical}, the center of mass is outside the supporting polygon but there is at least one leg in stance phase on the left and the right side (Fig. \ref{fig:example_pre}c.2). 
Despite not being statically stable, the statically unstable configurations can be made dynamically stable when the speed increases \cite{hildebrand1980adaptive} or when combined with a low-level controller \cite{kalakrishnan2010fast,park2017high,bellicoso2018dynamic}. In other words, the loss of static stability could be compensated by the acquired dynamic stability. In the unstable configurations, also known as unstable lateral-couplet gaits \cite{hildebrand1967symmetrical}, either the left or the right side of the legs are all in swing phase (Fig. \ref{fig:example_pre}c.3), which makes it more difficult to stabilize\footnote{in the case of limbless sidewinding, unstable configurations are defined as those with no contact, see Fig. S4}.
We define the static stability metric as the fraction of the gait cycle spent in statically stable configurations.
Note that this measure only applies to the gaits with statically stable and statically unstable configurations; the appearance of unstable configurations will contradict our assumptions. Therefore, we define the measure of static stability to be 0 if there exists unstable configurations in the gait.

We numerically calculated the static stability for the quadrupedal, hexapod, and myriapod systems in Fig. \ref{fig:sim}. 
As one might expect, when comparing the same gait parameters (duty factor $D$ and phase lag $\Phi_{lat}$) among different systems, the static stability increases with the number of legs.
Similarly, an increase in duty factor results in an increase in static stability.
Moreover, we observe that the diagonal sequence ($\Phi_{lat}>0.5$) is in general less stable than the lateral sequence ($\Phi_{lat}<0.5$).
Thus, most diagonal sequence gaits are stable only for systems with a greater number of legs, such as myriapods.
The static stability is also strongly correlated with the lateral phase lag. 
Specifically, gaits are more statically stable as $|\Phi_{lat}-0.5|\rightarrow0$ and they become ``closer'' to an alternating tripod gait.

Surprisingly, modulating the lateral phase lag only affects the static stability, while body speed is not correlated with the lateral phase lag. 
On the other hand, animals including myriapods \cite{manton1952evolution} and quadrupedal lizards \cite{farley1997mechanics,reilly1997a,reilly1997b} have been observed to modulate the lateral phase lag as speed increases. 
In other words, in biological systems, the loss of static stability is compensated by a gain in speed while our findings indicate that speed is independent of lateral phase lag modulation. We hypothesize that this discrepancy is due to differences in whole-body coordination, which we consider in the next section.

\subsection*{Experimental Results on Speed and Stability}

\begin{figure*}
    \centering
    \includegraphics[width=0.8\linewidth]{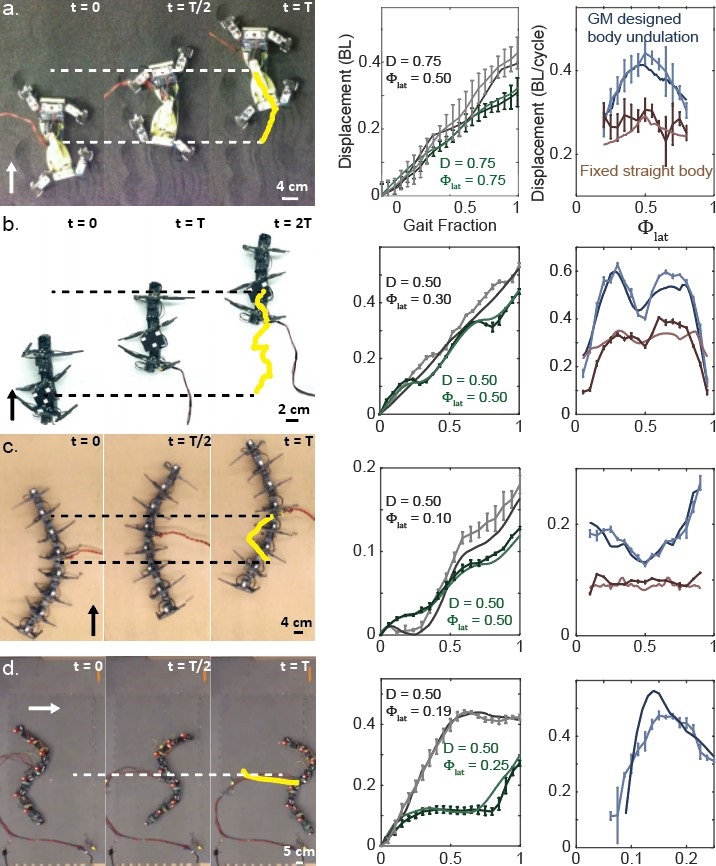}
    \caption{\textbf{Verification of the theoretically generated gaits in the robotic models} (left column) Gait cycle of each robot (a: quadruped, $D = 0.75$ and $\Phi_{lat} = 0.5$; b: hexapod, $D = 0.5$ and $\Phi_{lat}=0.3$; c: myriapod, $D = 0.5$, $\Phi_{lat} = 0.1$; d: sidewinder). The arrows show the direction of locomotion and $T$ is one gait cycle.  The center of mass trajectories (yellow) are given in the last snapshots. (Middle column) The comparison of simulations (solid curves) and experimental data (curves with error bar) of displacement over time for each system. Two gaits with body undulation coordinated with geometric mechanics (GM) are illustrated for each system. (Right column) The relationship between the lateral phase lag, $\Phi_{lat}$, and the displacement for the same system either with fixed straight backbone (red) or with coordinated lateral body undulation (blue). The color scheme and axes in (b, c, d) is the same as in (a).}
    \label{fig:exp}
\end{figure*}

Using robophysical models, we tested the locomotor performance of gaits with a range of lateral phase lag for quadrupedal, hexapod, myriapod, and sidewinder systems (Fig. \ref{fig:exp}). 
The quadrupedal experiments were performed on granular media (poppy seeds); other experiments were performed on hard ground.
The duty factor for the hexapod, myriapod and the sidewinder systems were fixed to $D = 0.5$, and the duty factor for the  quadrupedal system was set to $D = 0.75$ for reference  (see the Materials and Methods for additional experiment details).  
Note that in this section, there is no body undulation in all gaits.

We measured gait speed via the number of body lengths traveled per gait cycle.
Interestingly, the range of gaits showing the most theory-experiment discrepancy overlaps with the range of gaits that are not statically stable.
Since our predictions are based on 2D physics calculations, they cannot capture 3D unstable behaviors, such as tipping over and falling to the ground. 
Therefore, we hypothesize that the discrepancy between our hexapod theory and experiments is caused by static instability. 
Note that our experiments on quadrupeds were performed on poppy seeds, where the ventral surface often was in contact with the environment.
In our myriapod experiments, configurations tend to be mostly statically stable given their large number of legs.
Therefore, the effect of static stability was only critical in our hexapod experiments.

\begin{figure*}
    \centering
    \includegraphics[width=0.8\linewidth]{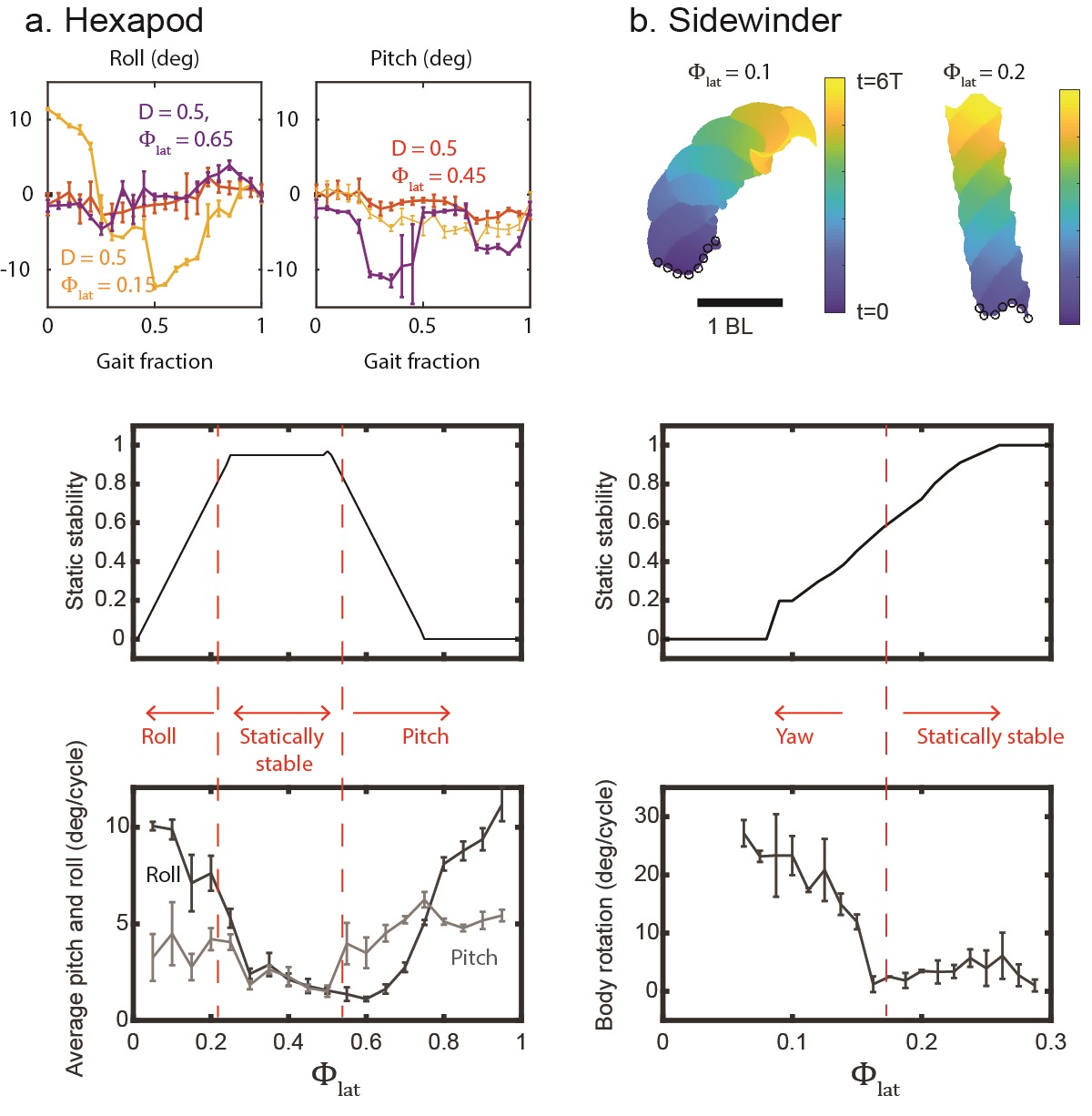}
    \caption{\textbf{The effect of static stability on locomotion performance.}  In the left column (a.), the body roll and pitch over the course of the hexapod experiments are recorded as a function of gait fraction. Three gaits ($D=0.5$, $\Phi_{lat}=0.65$ in purple; $D=0.5$, $\Phi_{lat}=0.45$ in red; and $D=0.5$, $\Phi_{lat}=0.15$ in yellow) in Hildebrand gait space are compared. In the middle row, we show the theoretical prediction of static stability as a function of lateral phase lag. In the bottom row, we show the average$\pm$SD experimental body roll and pitch as a function of the lateral phase lag. In the right column, (b.), a similar analysis is performed for the sidewinder experiments. The top-right shows the trajectory of body motion over six gait cycles, where the color scale represents the evolution of time. We marked the initial position of the robot in the black circles. In the middle row, we showed the theoretical prediction of static stability as a function of lateral phase lag. In the bottom panel of Fig 7b, The body yaw angle is recorded as a function of lateral phase lag. }
    \label{fig:error}
\end{figure*}

To test our hypothesis that static stability is the source of the theory-experiment discrepancy, we characterized unstable behaviors by the roll and pitch of the robots. 
We recorded the body pitch and roll during the course of the robophysical hexapod gaits.
The experimental data for these experiments over three gait settings ($D=0.5\ \Phi_{lat}=0.15$, $D=0.5\ \Phi_{lat}=0.45$, $D=0.5\ \Phi_{lat}=0.65$) are compared in Fig. \ref{fig:error}a.
We observed that only the statically stable hexapod gait ($\Phi_{lat}=0.45$) has both low pitch and low roll. 
The unstable hexapod gaits have either high roll angle ($\Phi_{lat}=0.15$) or high pitch angle ($\Phi_{lat}=0.65$). 
We calculated the average pitch and roll for each gait, and compared them with the numerical predictions of static stability. 
We observe that the range of low average pitch and roll overlaps with the range of statically stable gaits.
When the hexapod body is in configurations with low roll and low pitch, the experimental data agree with the theoretical predictions. 

\section*{Body-leg Coordination in Hildebrand Gait Formulation}

\subsection*{Geometric Mechanics to Coordinate Lateral Body Undulation}

As discussed in the previous sections, speed is not correlated with the the lateral phase lag when there is no body undulation. 
However, previous experimental gait studies with lizards and myriapods \cite{manton1952evolution,farley1997mechanics} have found that modulation of lateral phase lag is associated with changes in the lateral body undulation.
For example, lizards increase the amplitude of their lateral body undulation during transitions from lateral sequence walking to trotting or even diagonal sequence gaits \cite{farley1997mechanics,sukhanov1974general,reilly1997a,reilly1997b,hutchinson2019divergent,white1994locomotor}.
Similarly, myriapods change their leg wave pattern (lateral phase lag) at high speed while simultaneously increasing lateral body undulation amplitude  \cite{manton1952evolution}.
Accordingly, we hypothesize that modulating the lateral phase lag can regulate the balance between speed and stability only if properly coordinated with lateral body undulation.

To account for these observations, we introduce the lateral body bending angle $\alpha$ (Fig. \ref{fig:prescribe}).
We prescribed lateral body undulation by propagating a wave along the backbone from head to tail \cite{hirose2009snake}:
\begin{align}
    \alpha(\phi_b,i)=A_\alpha \text{cos}(\phi_b - 2\pi (i-1) \Phi_{lat}^b)
    \label{eq:latBodyUndu}
\end{align}

\noindent where $\alpha(\phi_b,i)$ is the angle of $i$-th body joint at phase $\phi_c$, $2\pi \Phi_{lat}^b$ is the phase lag between consecutive joints. For simplicity, we assume that the spatial frequency of the body undulation wave and the contact pattern wave are the same, i.e. $\Phi_{lat}^b = \Phi_{lat}$. 
The body shape can then be described as the phase of contact, $\phi_c$, and the phase of lateral body undulation $\phi_b$. 
These two independent phase variables represent a reduced shape space (see Materials and Methods) on a two-dimensional torus on which we can apply geometric mechanics gait design techniques to optimize body-limb coordination (Materials and Methods).

The geometric mechanics gait design framework separates the configuration space of a system into two spaces: the position space and the shape space.
The position space represents the location (position and orientation) of a system relative to the world frame, while the shape space represents the internal shape (joint angles) of the system.
The geometric mechanics framework then establishes a functional relationship to map velocities in the shape space into velocities in the position space; this functional relationship is often called a \emph{local connection}. The curl of the local connection, which we call a ``height function'' can then be used to design, analyze, and optimize gaits.

Using geometric mechanics tools \cite{hatton2015nonconservativity,toruspaper,Marsden}, we derived height functions and designed gaits (Materials and Methods). 
Fig. \ref{fig:example_pre} and Fig. \ref{fig:example_evaluate} show examples of coordination between the lateral body undulation and contact phase derived with geometric mechanics. 
We also provided an example of coordinating the body undulation and contact pattern for sidewinding in Fig. S4.
Once we designed a coordination pattern $\phi_c \rightarrow \phi_b$ in the reduced shape space, we can convert that pattern into both a contact pattern and body undulation.

\begin{figure}
    \centering
    \includegraphics[width=0.5\linewidth]{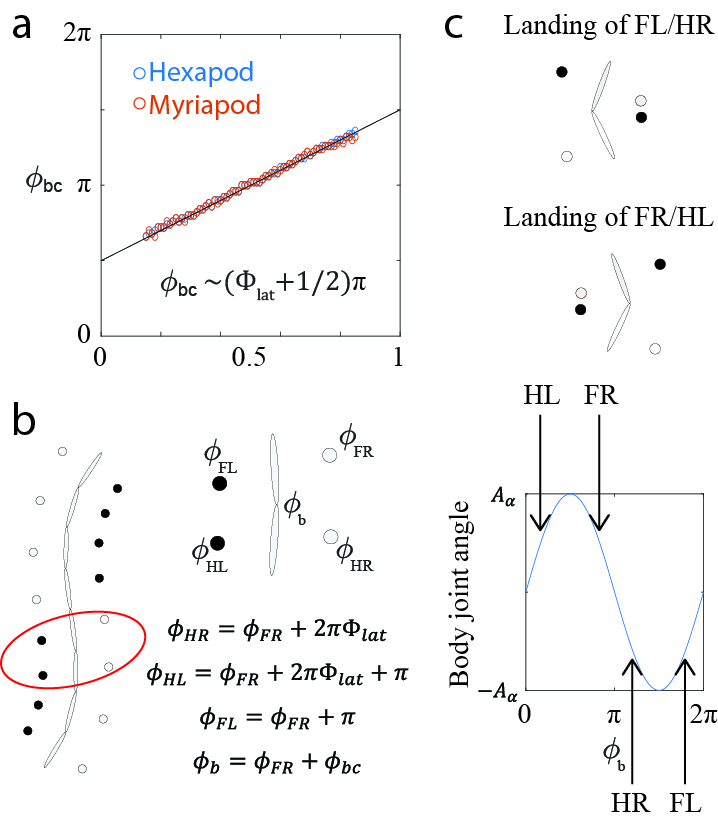}
    \caption{\textbf{Physical intuition in body-leg coordination} (a) The relationship between $\Phi_{lat}$, lateral phase lag, and $\phi_{bc}$, the optimal phasing between body and leg. $\phi_{bc}$ is numerically calculated from a height function (Fig. \ref{fig:example_pre}). The empirical data for the hexapod (blue circle) and myriapod (red circle) are compared. (b) Consider a quadrupedal ``sub-unit'' consisting of two pairs of legs and one body-joint. The Hildebrand prescription allows us to write the phase relation of each leg and the body bending with respect to the fore right leg (FR). (c) To maximize locomotive performance with body-bending, at FL (fore left) and HR (hind right) touchdown, the body is bent clockwise; and at FR (fore right) and HL (hind left) touchdown, the body is bent counterclockwise \cite{chong2021coordination}. Given this empirical relation $\phi_{bc}\sim(\Phi_{lat}+1/2)\pi$, the HL/FR and HR/FL touchdown phases are symmetrically distributed around the peaks of the bending trajectory, which we use to coordinate body-bending with foot contacts.}
    \label{fig:physicaIntui}
\end{figure}

We quantified the body-leg coordination by its phase lag: $\phi_{bc}: \phi_c - \phi_b$. Interestingly, we observed that the empirically calculated $\phi_{bc}$ has a linear relationship with $\Phi_{lat}$ (Fig. \ref{fig:physicaIntui}). We next seek to investigate the physical intuition behind this relationship. We first decomposed the body-leg coordination to a single ``sub-unit,'' which we define as two pairs of legs and one body joint. Our Hildebrand-based approach then allows us to prescribe the phase of each feet and the body bending. Previous work \cite{chong2021coordination} found that in effective gaits, when the HL/FR feet land, the body is bent clockwise, and when the HR/FL feet land, the body is bent counterclockwise. We can encode this relation by $\phi_{bc}\sim(\Phi_{lat}+1/2)\pi$. Then, the FR and HL foot touch-down is symmetrically distributed around the peak of the clockwise body bending angle, and the touch-down of FL and HR feet are symmetrically distributed around the peak of counterclockwise body bending angle. Via this relationship, we posit that despite the seemingly complicated whole-body motion, the optimal body-leg coordination is achieved by locally coordinating each sub-unit of two legs and a body joint.

\subsection*{Numerical Prediction of Speed and Stability}

We used a numerical simulation to predict the gait speed and stability at a range of lateral phase lags and duty factors for the quadrupedal, hexapod, myriapod and sidewinder systems. 
We observed that modulating the lateral phase lag can regulate the balance between speed and stability if properly coordinated with lateral body undulation. 
The loss of static stability is compensated by a gain in speed only when the body and limb phases are properly coordinated.
These observations were derived by plotting gait speed and stability against the extended Hildebrand gait parameters, shown in Fig. \ref{fig:sim}.
The addition of body undulation changes slightly changes the static stability, as depicted in Fig. S1.

\subsection*{Experimental Results}

We tested the locomotion performance of systems with discrete contact and coordinated lateral body undulation using robophysical models (Materials and Methods for details). 
We recorded the displacement over time for two gaits in each system (Fig. \ref{fig:exp}). Our numerical predictions quantitatively agree with experiments not only in the average displacement per gait cycle, but also in the time evolution of the displacement.

The only notable theory-experiment discrepancies occur in the hexapod and the sidewinder systems. As discussed earlier, static instability can lead to theory-experiment discrepancy for hexapods and sidewinders due to the planar assumptions made in our theoretical model. 
To investigate this discrepancy further, we studied the effect of static instability on sidewinders and observed that some gaits result in significant yaw (Fig. \ref{fig:error}b), such that the robot's path deviates from the desired straight-line course.
Comparing the net yaw change per gait cycle with the numerical predictions of static stability reveals that significant yaw only occurred in gaits with low static stability. As static stability increases (for sidewinding, stability increases with the lateral phase lag), the unmodelled turning vanished. 

\section*{Body-leg Coordination in Biological Locomotors}

\begin{figure}
    \centering
    \includegraphics[width=0.8\linewidth]{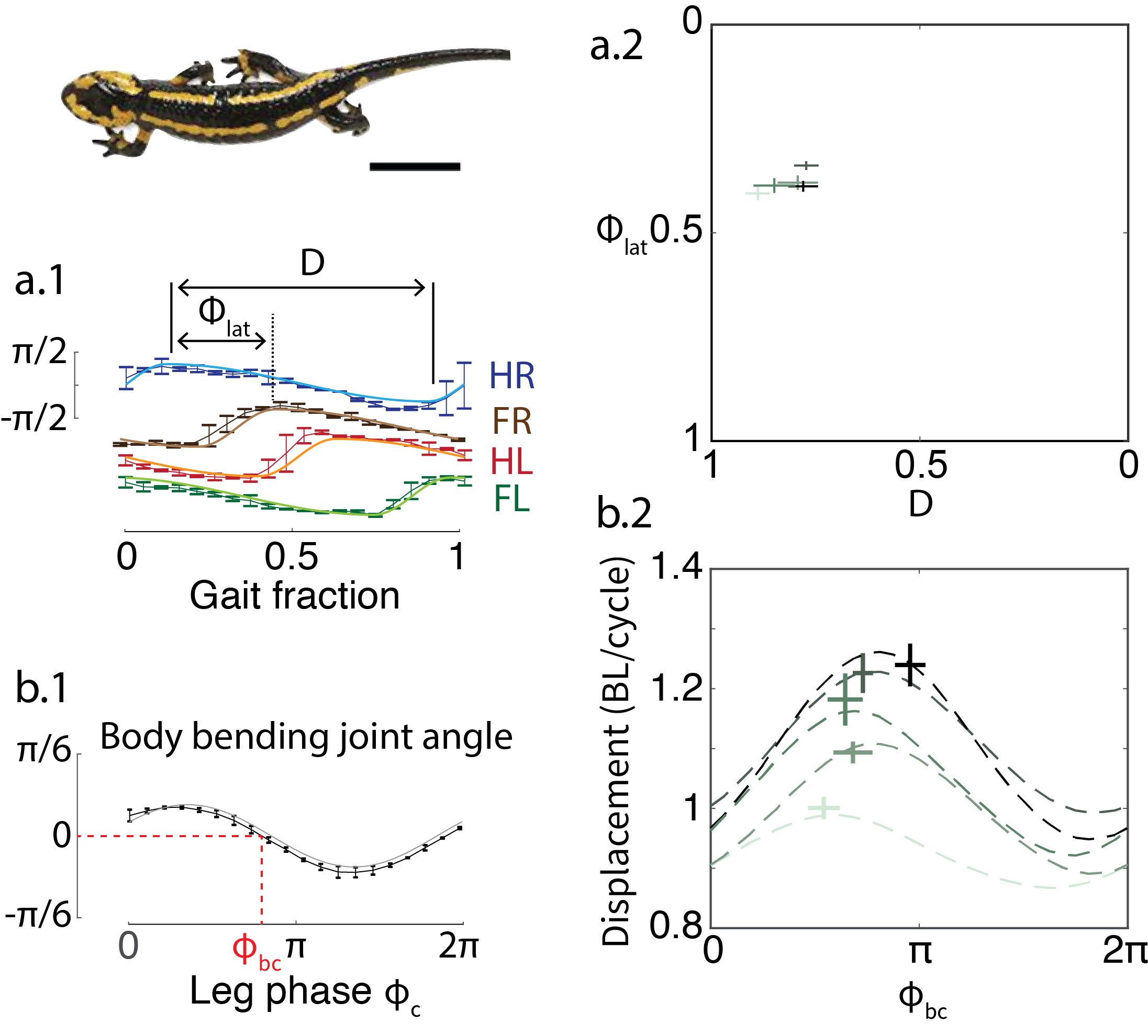}
    \caption{\textbf{Analysis of salamander (\textit{Salamandra salamandra}) locomotion using the Hildebrand framework and geometric mechanics} (a.1) Estimation of the duty factor, $D$, and lateral phase lag, $\Phi_{lat}$ from animal joint angle trajectories. Curves with error bars are the average leg shoulder (hip) angle over three cycles. The lighter-color solid curves are piece-wise linear sinusoidal functions (defined in Eq. \ref{eq:legmove}) fit to the tracked data. (a.2) Estimated $D$ and $\Phi_{lat}$ for animal locomotion under different speeds. (b.1) Estimating $\phi_{bc}$ from body bending angle trajectories. (b.2) Relationship between $\phi_{bc}$ and speed, measured in body lengths (BL) per cycle. The prediction made with geometric mechanics is shown as dashed curves. The measured salamander data are presented by crosses in the same color as their corresponding prediction curves, where the length and height of the crosses denote the standard deviation of the measured animal data. Scale bar near salamander photo indicates 30mm.}
    \label{fig:biocompare1}
\end{figure}

Symmetric gaits in quadrupedal animals can be categorized using Hildebrand analysis~\cite{hildebrand1965symmetrical,hildebrand1967symmetrical}. Recent work showed that a geometric mechanics framework can predict the optimal body-leg coordination for fire salamanders (\textit{Salamandra salamandra})~\cite{zhong2018coordination,chong2021coordination}. However, the means by which salamander modify their leg movements and body-leg coordination in response to speed changes was previously unstudied. In this work, we recorded fire salamanders moving on sand. Five individuals were recorded, and their foot placement and backbone positions tracked. From the tracking data we measured gait parameters such as duty factor, lateral phase lag, amplitude of body bending, and amplitude of leg movements. We then used geometric mechanics to predict the optimal body-leg coordination for salamanders walking at various speeds. We observed quantitative agreement between the geometric mechanics prediction and the measured biological data (Fig. \ref{fig:biocompare1}).

\begin{figure}
    \centering
    \includegraphics[width=0.8\linewidth]{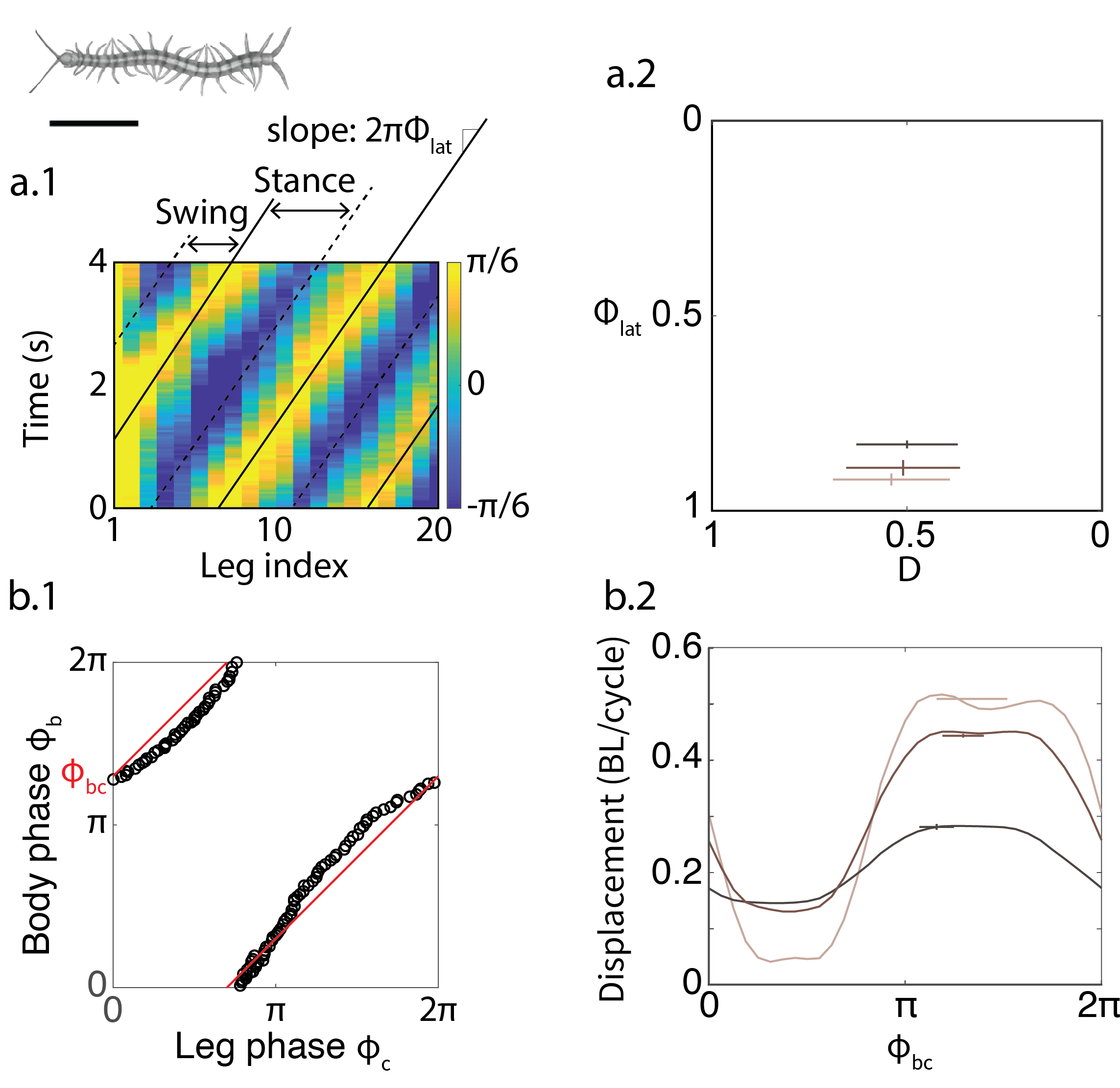}
    \caption{\textbf{Analysis of centipede (\textit{Scolopendra polymorpha}) locomotion using the Hildebrand framework and geometric mechanics} (a.1) Estimation of the duty factor, $D$, and lateral phase lag, $\Phi_{lat}$ from animal joint trajectories. The colorbar here denotes the shoulder joint angle for each leg on right-hand side. (a.2) Estimated gait parameters $D$ and $\Phi_{lat}$ for the centipede's locomotion. (b.1) Estimating $\phi_{bc}$ from body phase and leg phase. (b.2) Relationship between $\phi_{bc}$ and the speed, measured in body lengths per cycle. The prediction made with geometric mechanics is shown as solid curves. The measured centipede data are presented by crosses in the same color as their corresponding prediction curves, where the length and height of the crosses denote the standard deviation of the measured animal data. Scale bar near centipede photo indicates 30mm.}
    \label{fig:biocompare2}
\end{figure}

Beyond quadrupedal animals, our methods can also be applied to study animals with various number of legs and backbone segments. Centipedes are known to be fast-moving locomotors: certain centipedes are the fastest-running terrestrial arthropods~\cite{manton1952evolution,anderson1995axial}. Given their high speeds, past work often used dynamic models to analyze their locomotion~\cite{aoi2013instability,yasui2019decoding}. We hypothesized that despite their high speeds, centipede locomotion can be analyzed with our quasi-static geometric model.
To test this hypothesis, we recorded videos of centipedes (\textit{Scolopendra polymorpha}) under different speeds. Three individuals were recorded, their leg and body positions tracked, and their gait parameters estimated. We then used geometric mechanics to predict the optimal body-leg coordination. We once again observed quantitative agreement between geometric mechanics predictions and the measured animal data (Fig. \ref{fig:biocompare2}).

\section*{Discussion and Conclusion}

\subsection*{Principles of gait modulation}

In this paper, we developed a general gait design framework for a broad class of locomotors: multi-legged robots (with an arbitrary number of pairs of legs) with an articulated backbone, including limbless sidewinding. Specifically, we  extended the Hildebrand gait formulation \cite{hildebrand1965symmetrical,hildebrand1967symmetrical}, originally used to categorize symmetric quadrupedal gaits, and combined it with modern geometric mechanics tools to investigate optimal leg-body coordination.
We showed that the symmetry in Hildebrand quadrupedal gaits is conserved for other locomotors: it is simple enough to enable physical interpretation of the gait parameters; on the other hand, it is sufficiently rich in content, offering a scheme to modulate gaits in a diversity of robot shapes.
These properties enable our framework to link well-studied locomoting systems like quadrupeds and hexapods with less-studied systems like myriapods, generating new opportunities to transfer insights among and compare between different locomoting morphologies. Given a new robot with arbitrary pairs of legs or without legs, our framework can immediately provide effective open-loop gaits, which can serve as the basis for closed-loop adaptive or data-driven/learning-based control algorithms. 

Our gait principles reveal insights into proper contact scheduling. These principles could serve as a starting point for additional layers within in a robot's control architecture or even for mechanical design iterations.
For example, \cite{ozkan2020systematic} found that while direct application of gait design tools can prove ineffective in rough terrain, adding passive leg compliance can greatly improve performance in this environment.
Our proposed framework can not only simplify the gait design and modulation process for robots with different morphologies in various homogeneous environments, but can also provide a guideline for designing robot controllers that navigate unstructured environments and overcome obstacles.  Our framework can also be used to test hypotheses and therefore give novel insights into the control principles behind gaits in biological systems.

Finally, our framework facilitated testing hypotheses about the role of body undulation in multi-legged systems.
These observations can act as guidelines in the control of a variety of legged robots.
For example, in RHex \cite{saranli2001rhex}, a hexapod with  flexible legs attached to a rigid body, the duty factor is the only tuning parameter that can regulate the balance between speed and stability.
In other cases, such as in \cite{hoffman2011myriapod}, a segmented robot with a flexible backbone and contralateral legs coupled to a straight line (and therefore, have a fixed duty factor), the lateral phase lag acts as the salient parameter to balance between speed and stability when properly coordinated with lateral body undulation.

\subsection*{Insights from robotics to biological systems}

We have also shown here that once two gait parameters (duty factor and lateral phase lag) are specified, the gait is readily prescribed and can then be analyzed with geometric tools. 
To explore gait tuning principles for locomoting systems, we quantitatively investigated the effect of modulating gait parameters on locomotor performance.
As shown in Fig. \ref{fig:exp}, we found that in robots with a fixed straight backbone, the displacement per gait cycle is nearly invariant to the changes in the \emph{lateral phase lag}, $\Phi_{lat}$. On the other hand, in gaits where body undulation is properly coordinated with leg motions, $\Phi_{lat}$ affects the displacement. This seemingly counter-intuitive observation can help us develop hypotheses about gait modulation principles.

In addition to these robotics applications, our proposed control principles can also offer explanatory power to some hypotheses about biological locomotion.
For example, biological myriapods (\emph{Chilopoda}) can be categorized into direct-wave myriapods \cite{manton1952evolution} and retrograde-wave myriapods \cite{manton1952evolution}).
Direct-wave myriapods propagate their leg contact wave from tail to head (corresponding to $\Phi_{lat}<0.5$ in our modified Hildebrand formulation) while retrograde-wave myriapods propagate their wave from head to tail ($\Phi_{lat}>0.5$) \cite{kuroda2014common}.
Interestingly, Manton \cite{manton1952evolution} showed that there is no significant lateral body undulation in direct-wave myriapods regardless of their speed; instead, the only significant gait modulation at high speed is a decrease in duty factor.
On the other hand, gait modulation in retrograde-wave myriapods is much more complicated: they not only decrease the duty factor, but also increase the lateral phase lag. More importantly, they exhibit characteristic lateral body undulation at high speeds \cite{manton1952evolution,anderson1995axial}.
This observation is consistent with the principles discovered via our gait analysis methods, where we found that tuning the lateral phase lag can only improve the speed if accompanied with properly coordinated lateral body undulation.
Moreover, the contribution from lateral body undulation is greater for retrograde-wave myriapod locomotion than for direct-wave myriapods. This may be one of the reasons behind the biological observation that lateral body undulation is only characteristic of retrograde-wave myriapods \cite{manton1952evolution,anderson1995axial}.

\section*{Materials and Methods}

\subsection*{Geometric Mechanics}

We used tools from geometric mechanics, the application of differential geometry concepts to rigid body mechanics, to design the coordination between body undulation and contact patterns. 
In this section, we provide a concise overview of the tools used to design the coordination patterns.
For a more detailed and comprehensive review, we refer readers to~\cite{toruspaper,Marsden,chong2021coordination}.

The geometric mechanics gait design framework separates the configuration space of a system into two spaces: the position space and the shape space.
The position space represents the location (position and orientation) of a system relative to the world frame, while the shape space represents the internal shape (joint angles) of the system.
The geometric mechanics framework then establishes a functional relationship to map velocities in the shape space into velocities in the position space; this functional relationship is often called a \emph{local connection}.


\subsubsection*{Reduced Equation of Motion}

In kinematic systems where frictional forces dominate inertial forces, the equations of motion can be approximated by:
\vspace{-0.2cm}
\begin{equation}
    \boldsymbol{\xi}=\boldsymbol{A(\Phi)\dot \Phi},
    \label{eq:EquationOfMotion1}
\end{equation}

\noindent where $\boldsymbol{\xi}=[\xi_x \ \xi_y \ \xi_\theta]^T\in g $ denotes the body velocity in the forward ($x$), lateral ($y$), and yaw ($\theta$) directions; $\boldsymbol{\Phi}$ denotes the internal shape variables. In this work, $\boldsymbol{\Phi} = [\phi_c\ \phi_b ]^T$, representing the contact phase and the lateral body undulation phase. $\boldsymbol{A(\Phi)}$ is the local connection matrix, which encodes environmental substrate interactions. 

\subsubsection*{Numerical Derivation of the Local Connection Matrix}

The local connection matrix $\boldsymbol{A}$ can be numerically derived via force and torque balances \cite{hatton2013geometric}. The force and torque balance equations require a model of the ground reaction forces (GRF), such as granular material interaction and ground friction.  We summarize the GRF formula for our four robots above. Further details on the local connection derivation can be found in the Supplementary Material.

\begin{table*}[]
    \centering
\begin{tabular}{ |l|l|l| } 
\hline
 \emph{Robot} &  \emph{GRF Formula} &  \emph{Reference}\\ 
 \hline
 Quadruped  & Poppy Seed RFT & \cite{mcinroe2016tail,chong2021coordination} \\ 
 \hline
 Hexapod  & Anisotropic Coulomb Friction &  \cite{walker2019set,transeth20083} \\ 
 \hline
 Myriapod & Anisotropic Coulomb Friction & \cite{walker2019set,transeth20083} \\
 \hline
 Sidewinder & Isotropic Coulomb Friction & \cite{chong2020,rieser2019geometric}\\ 
 \hline
\end{tabular}
\caption{The ground reaction force formulas used to model quadrupeds, hexapods, myriapods, and sidewinders robophysical systems.}    \label{tab:my_label}
\end{table*}

\subsubsection*{Connection Vector Fields and Height Functions}

Once we obtain the local connection matrix, we can further analyze the system kinematics during locomotion. Each row of the local connection matrix $\boldsymbol{A}$ corresponds to a component direction of the body velocity. Each row of the local connection matrix over the shape space then forms a connection vector field.
Then, the body velocity can be computed via the dot product of connection vector fields and the shape velocity $\boldsymbol{\dot \Phi}$. 

A periodic gait can be represented as a closed curve in the shape space. The displacement resulting from a gait, $\partial \chi$, can be approximated by:
\vspace{-0.2cm}
\begin{equation}
    \begin{pmatrix} 
        \Delta x \\
        \Delta y \\
        \Delta \theta 
    \end{pmatrix}
    = \int_{\partial \chi} \boldsymbol{A(\Phi)\boldsymbol{d}\Phi}.
\label{eq:lineintegral}    
\end{equation}

According to Stokes' Theorem, the line integral along a closed curve $\partial \chi$ is equal to the surface integral of the curl of $\boldsymbol{A(\Phi)}$ over the surface enclosed by $\partial \chi$:
\vspace{-0.2cm}
\begin{equation}
    \int_{\partial \chi} \boldsymbol{A(\Phi)d\Phi}=\iint_{\chi} \boldsymbol{\nabla\times A(\Phi)}\boldsymbol{d}\phi_c\boldsymbol{d}\phi_b,
\label{eq:stokes}
\end{equation}

\noindent where $\chi$ denotes the surface enclosed by $\partial \chi$. 
The curl of the connection vector field, $\boldsymbol{\nabla\times A(\Phi)}$, is referred to as the \textit{height function}. 
The three rows of the vector field $\boldsymbol{A(\Phi)}$ can thus produce three height functions in the forward, lateral and rotational direction, respectively. 

The height function derivation simplifies the gait design problem to drawing a closed path in a Euclidean shape space. 
The body displacement from a path can be approximated by the integral of the surface enclosed by that path. 

\subsubsection*{Toroidal Shape Spaces}

In our gait prescription, the two shape variables are parameterized as cyclic phases, resulting in a toroidal shape space  $(T^2)$~\cite{kobayashi1963foundations}. Examples of height functions on toroidal shape spaces are shown in Fig. \ref{fig:example_pre}b.
The shape variables $\boldsymbol{\Phi}=[\phi_c, \ \phi_b]^T \in T^2$ correspond to the phase of contact and the phase of the lateral body undulation, respectively. 
A gait is a closed curve in the toroidal shape space (solid purple curve Fig. \ref{fig:example_pre}b), but as it is a non-Euclidean space, there is no clear ``surface'' enclosed by the path. 

To form an enclosed surface, Gong et al., \cite{toruspaper} introduced the notion of ``assistive lines'' in the Euclidean parameterization of the toroidal shape space. This method allows a surface integral to be calculated. In Fig. \ref{fig:example_pre}b, the surface integral is the area within solid lines is subtracted from the area of the surface enclosed in the upper left corner.
For simplicity, in our optimization we assumed that the mapping between the two phase variables is linear and that the body and legs share the same temporal frequency, i.e., $\phi_b = \partial \chi (\phi_c) = \phi_c + \phi_0$, where $\phi_0$ is the phase offset between lateral body undulation and contact pattern to be optimized.

\subsection*{Simulation}

We performed a numerical simulation to predict locomotive performance, and compared these results to those obtained from robophysical experiments. 
Specifically, we prescribed the contact state and the joint angle of each leg by a single variable, $\phi_c$, using Eq. \ref{eq:general_hild} and Eq. \ref{eq:legmove}.
Similarly, we prescribed the lateral body undulation by another variable, $\phi_b$, using Eq. \ref{eq:latBodyUndu}. The amplitudes of leg and body joint angles are listed below.

\begin{center}
\begin{tabular}{ |c|c|c| } 
\hline
 \textbf{Robot} &  $A_\theta$  & $A_\alpha$ \\ 
 \hline
 Quadruped & $30^\circ$ & $30^\circ$\\ 
 \hline
 Hexapod  & $10^\circ$ & $10^\circ$\\ 
 \hline
 Myriapod & $12^\circ$ & $17^\circ$\\ 
 \hline
 Sidewinder & N/A  & $5.6L$  (rad)\\ 
 \hline
\end{tabular}
\end{center}

Note that the amplitude of sidewinder body undulation is related to the lateral phase lag, such that the  ``relative curvature,'' that is, the maximum curvature of the backbone of limbless locomotors \cite{astley2020surprising,sharpe2015locomotor,rieser2019geometric}, remains constant.

Given the body-limb coordination function $\partial \chi: \phi_c \rightarrow \phi_b$ as described above, the shape variable $\boldsymbol{\Phi}$ and shape velocity $\boldsymbol{\dot \Phi}$ can be rewritten as:

\begin{align}
      \boldsymbol{\Phi} =     \begin{bmatrix} 
        \phi_c  \\
        \partial \chi (\phi_c)  
    \end{bmatrix},  \ 
    \dot{\boldsymbol{\Phi}}=     \begin{bmatrix} 
        1   \\
        \frac{\boldsymbol{d} \partial \chi (\phi_c)}{\boldsymbol{d}\phi_c}  
    \end{bmatrix} \dot \phi_c
\end{align}

The body displacement is computed by integrating the following ordinary differential equation \cite{hatton2015nonconservativity}:
\begin{align}
        g(t) &=  \int_{0}^{t} T_e L_{g(\phi_c)} A(\boldsymbol{\Phi}) \boldsymbol{d}{\boldsymbol{\Phi}} \\
        &=\int_{0}^{t} T_e L_{g(\phi_c)} A(\begin{bmatrix} 
        \phi_c  \\
        \partial \chi (\phi_c)  
    \end{bmatrix}) \begin{bmatrix} 
        1   \\
        \frac{\boldsymbol{d} \partial \chi (\phi_c)}{\boldsymbol{d}\phi_c}  
    \end{bmatrix}\boldsymbol{d}\phi_c, 
\end{align}

\noindent where $g= (x, y, \alpha)\in SE(2)$ represents the body frame position and rotation \cite{murray2017mathematical}. Note that $T_eL_{g}$ is the left lifted action with respect to the coordinates of $g$:

\begin{equation}
    T_eL_{g}=\begin{bmatrix} 
        \cos(\alpha) & -\sin(\alpha) & 0  \\
        \sin(\alpha) & \cos(\alpha) & 0  \\
        0 & 0 & 1
    \end{bmatrix}
\end{equation}

Integrating the ordinary differential equation throughout one period (from $t=0$ to $t=2\pi$), results in the body trajectory, from which we can determine the predicted displacements in the forward, lateral, and rotational directions over one gait cycle. Note that we neglect any inertial effects in this simulation.

\subsection*{Robophysical Experiments}

\paragraph{Robotic Models}
All of the robophysical models were designed in Solidworks and printed using Stratasys Dimension Elite 3D Printer. They are powered with an external power supply (12 V, 5A) and controlled via the MATLAB DYNAMIXEL SDK, interfacing with the servo motors through a Robotis USB2Dynamixel controller. All the robots have open-loop control such that gait parameters
are not changed during an experiment, and the control signals (servo positions) continue to be sent as a function of time, regardless of external forces or the tracking accuracy of the servos. \par 

The quadruped robot (Fig.S4a, 450 g., $\sim$40 cm long, \cite{ozkanAydin_LivingMach}) has four legs and an actuated trunk. Each limb is actuated with two Dynamixel XL-320 servos (stall torque 0.39 [N.m]) to control the vertical position and the step size of the leg (45 mm-high steps, Fig.S4a).  The body joint servo (Dynamixel AX-12) controls the horizontal bending.  The legs have a cube shape with $24\times24$ mm$^2$ surface area. \par

The hexapod robot (Fig.S4b, 300 g., $\sim$25 cm long) has a segmented body (three segments) with pairs of legs in each segment. The vertical and horizontal motion of the legs in a segment are coupled (out-of-phase) and controlled by two Dynamixel XL-320 servos (Fig.S4b). The body joint servos (Dynamixel XL-320) controls the horizontal bending of the segments.  The legs have pointed feet.  \par

The myriapod robot (Fig.S4c, $\sim$1000 g., $\sim$72 cm long) has a eight body segments, similar to the segments of the hexapod robot \cite{OzkanAydinRobosoft2020}. Each segment has a pair of rigidly connected legs 12 cm in length. There are three servos (Dynamixel XL-320) in each segment; one controls horizontal body bending and two control the fore/aft and up/down motion of the legs (Fig.S4b).  

The sidewinder robot has seven segments, each of which contains two joints connected at an angle of 90$^o$ (Fig.S4c). Each joint is comprised of a AX-12 servo motors (stall torque = 1.5 [N.m]). The horizontal motors vary the lateral wave, and the vertical motors create a changing contact pattern.

\paragraph{Experimental Setups and Data Analysis}
We used an Optitrack motion capture system (including 4-6 Naturalpoint, Flex13 cameras, 120 fps and Motive software) to capture the position and orientation of the reflective markers attached to the robots. The data was analyzed in Matlab. \par

Quadruped robot experiments were performed on a trackway filled with $\sim$1 mm diameter poppy seeds \cite{ozkanAydin_LivingMach}. Before each experiment we fluidized the bed using four vacuums to prepare a uniform loosely packed state. Each experimental condition, consisting of the quadruped robot and a set of gait parameters was repeated three times for a total of nine gait cycles. \par 

The hexapod and myriapod robot experiments were performed on a cardboard and particle board surface, respectively. Before each experiment, the joints were set to their neutral positions. The robots were allowed to run for three cycles (five trials/gait).

Sidewinder experiments were performed on a foam mat surface to reduce slip. Each experiment was started from the same position and repeated three times, for five to six gait cycles per trial. 

\paragraph{Salamander data analysis}

In salamander experiments, individual animals walked along a straight trackway filled with 300-$\mu$m glass particles. 
Three cameras (GoPro Hero3+, 720 pixel resolution) were positioned around the trackway and recorded synchronized videos at $120$~FPS. 
All experiments were approved by the Royal Veterinary College's Clinical Research Ethical Review Board, approval number 2015 1336. Salamanders were captured under collection permit \# 2016/001092 provided by the Government of the Principality of Asturias. No animals were harmed for the experiments, and animals had rest periods in between data collection trials. Experiments were conducted in a humidity-controlled laboratory at the University of Oviedo, Spain. The temperature (~18°C) and light cycle (12hr dark, 12hr light) were maintained at constant levels.

At least three gait periods were recorded in each experiment. Limb positions, body angles, and footfall timing are manually extracted from each recording. We fitted the animal body angles with the first two terms of Fourier Series as in \cite{chong2021coordination}.

\paragraph{Centipede Data Analysis}

In centipede experiments, individual animals ran along a flat, hard trackway. One overhead camera were was to record centipede locomotion. Positional data were extracted from videos with animal pose estimation software DeepLabCut (DLC) \cite{nath2019using}. Ten frames from each video were extracted and manually labeled. DLC would then provide positions for labeled points on all of the other frames. The positions of feet and body segments were labeled.

\section*{References}
\bibliography{iopart-num}

\clearpage 

\clearpage 

\clearpage

\clearpage 

\clearpage 

\clearpage

\clearpage 

\clearpage 

\end{document}